\documentclass[11pt]{article}

\usepackage[utf8]{inputenc}
\usepackage[T1]{fontenc}
\usepackage[english]{babel}
\usepackage{amsmath,amssymb,amsfonts}
\usepackage{graphicx}
\usepackage{geometry}
\usepackage{booktabs}
\usepackage{array}
\usepackage{hyperref}
\hypersetup{
    colorlinks=true,
    linkcolor=blue,
    citecolor=blue,
    urlcolor=blue
}
\usepackage{caption}
\usepackage{float}

\title{\textbf{A Method for Layer Bit-Width Allocation in LLM Quantization via Performance Maximization Under a Quality-Degradation Constraint}}

\author{A.K. Safronov \\ \small safro@sfedu.ru}
\date{}

\begin{document}

\maketitle

\begin{abstract}
This paper proposes a layer bit allocation method for Gemma-3-1B, which formulated the problem as a performance maximization (latency decrease) function under a given degradation budget constraint (the allowable level of generation quality loss). The allocation differs from time- and money-consuming uniform layer quantization methods such as GPTQ or AWQ and allocation methods without performance acceleration proof such as MixLLM or TorchAO. To do this, the author used the layer sensitivity profile obtained in the previous research (SA-PTQ) and implemented it through activation pass-through mode in TensorRT LLM. The precision of each layer is individually determined in blocks according to the previously introduced grouping (5+5, 10+10, all26), thus separating the contribution of FFN, Attention, and lm\_head to the overall speedup.

The clock speed measurements were made in 13 W8A8 variants on the RTX 5090. As a result, it was found that for FFN and lm\_head, the time costs of quantization/dequantization procedures are compensated by the introduction of integer arithmetic. In the case of short context lengths, the reverse situation is observed for Attention: the addition of another quantization step leads to slower model execution. Besides, the manual implementation of SmoothQuant for TensorRT LLM was proposed due to the export process’s failures, which is not available for lm\_head.

The highest performing solution under the joint consideration of all three criteria with minimal degradation was FFN 5+5 configuration with lm\_head, which demonstrated 11.0\% latency reduction with negligible quality loss (98.90\% Top-1 agreement, +0.85\% perplexity degradation). However, a significant increase in performance by 19.1\% was possible under acceptable quality loss conditions for FFN all26 + lm\_head configuration. Several optimizations were suggested to further accelerate the model: fused INT8 attention kernels, KV-cache quantization, FP8 conversion instead of INT8, and partial Attention quantization similar to the FFN block.
\end{abstract}


\section{Introduction}

The recent boom of distillation methods gave rise to the appearance of compact and versatile small language models (sLLMs) which can run on the consumer hardware. Gemma 3 1B is one such model --- it uses about 2 GB of video memory, which makes it a viable candidate for the user-facing applications, including the video games requiring real-time generation.

The standard way of quantizing such models with the state-of-the-art GPTQ~\cite{ref2} or AWQ~\cite{ref3} quantization methods is to apply the same set of weights to all the layers indiscriminately. However, as it was shown by the author in "A Layer Importance Metric for Quantization...", the sensitivity of the model layers to quantization is vastly different --- with the FFN block having a normalized sensitivity of $Q=0.796$ (mean SQNR of 29.70 dB) and the embedding matrix in the tokenizer being surprisingly robust with a value of $Q=0.819$ (mean SQNR of 45.61 dB), and other layers staying in between. This effectively creates groups of layers which can have much higher compression ratios than others.

sLLM class models' arrival made the question of why to use the mixed-precision quantization with varying ratios for different layers even more intriguing, as the memory savings are no longer critical for the 8 GB VRAM GPUs and above: a 2 GB model is easily placed in such a GPU with no memory issues. The issue of the computational efficiency still stands, and the only existing work which attempted to address it, MixLLM~\cite{ref5} and TorchAO~\cite{ref6} , does not provide the evidence of any specific layers or operations within the modern transformer architecture benefitting from the mixed-precision nature of their approach more than others and, by extension, justifying the additional overhead from the per-layer quantize/dequantize steps at the model boundaries.

The present paper introduces the method for performing the selective mixed-precision quantization based on the SA-PTQ sensitivity estimates and activation pass-through mode in TensorRT-LLM~\cite{ref7} and provides the concrete evidence for the benefits for such an approach in terms of FLOPs per second for the model with the per-layer ratios, based on the real-world measurements with TensorRT-LLM. Contrary to the claims of the methods which effectively remove the per-layer quantize/dequantize overhead by advertising it as a general property of mixed-precision weights, the present paper actually measures the said overhead and its impact on the per-layer FLOPs for various operations in the transformer, demonstrating that for the FFN and lm\_head layers the overhead is always offset by the gains from performing the calculations in the INT8 domain while for the short context lengths Attention operations, on the contrary, actually lose performance due to the overhead. The method allows to accelerate the inference by up to 11\\% with no loss in the generation quality (FFN 5+5 + lm-head) and up to 19\\% with the acceptable quality loss (all26) according to the wall-clock measurements on TensorRT-LLM.

\textbf{The goal of this work} is to apply the SA-PTQ layer importance metric developed in the author's previous work to the practical task of mixed-precision quantization of the Gemma-3-1B language model using TensorRT-LLM. The work provides an empirical assessment of the trade-off between inference speedup and generation-quality degradation for three key model subsystems -- FFN, the Attention mechanism, and lm\_head -- individually and in combination, with the aim of identifying quantization configurations that provide the maximum hardware gain under controlled information loss.

\subsection{Domain analysis}

The availability of distillation and compact-training approaches has led to the recent emergence of sLLM-class language models (LLM) capable of being run on commodity PC GPUs. Such models involve a small number of parameters (~10$^{9}$-$10^{10}$), which allows their deployment in model-inference applications, including video games. One such example is Gemma 3 1B~\cite{ref8}, which takes about 2 GB of VRAM in FP16 precision, thus making it deployable in a user-end application as a game engine. This particular model was selected for this work due to its relative availability of implementation details, which helps guide the experiments and discussion. It should be noted that the presented approach is applicable to any similar, compact LLM.

Quantization~\cite{ref19} is a technique which allows reducing the bit-width of weights and activations of a neural network, e.g. from FP16 to INT8, thus reducing the memory footprint and accelerating the calculations due to the more efficient integer arithmetic in modern GPUs. For language models, quantization techniques are mostly applied at the inference stage, where the model weights are already trained and do not benefit from further optimization~\cite{ref12}.

Modern transformer-based language models are complex structures which consist of dozens of blocks of different types. The most common types are multi-head attention blocks and FFN blocks, as well as the word embedding matrix~\cite{ref1}. The work by the authors which is included as~\cite{ref1}, has reported that the different blocks and layers within a block are not equal in their sensitivity to quantization~--~in particular, the FFN block has a normalized sensitivity score of about $Q=0.796$, while the embedding matrix has a much lower sensitivity of only about $Q=0.819$~ .

This observation allows creating a sensitivity profile of the model being quantized, where the different layers of the model have different per-sample sensitivity scores. It is this profile that is used as the basis for the approach described in this paper: the sensitivity score of each layer is used as a proxy for its quality at a particular bit-width, thus defining the maximal bit-width (among 5+5/10+10/all26 variants) that can be used for this layer. This allows picking the parts of the model which are the least sensitive to quantization, and experimenting with higher bit-widths for them to accelerate the computation without loss of quality.

The mixed-precision quantization approach introduces additional complexity due to the necessity to cast the activation tensors between the blocks of different precision at the forward/inference pass. Due to the different representations, each such transition requires additional de-quantization/re-quantization steps which add overhead and reduce the benefits of mixed-precision inference. As will be shown in this work, this complication has not been addressed in previous works, and constitutes the main problem studied in this paper.

\subsection{Related work}

A number of post-training quantization methods for language models currently exist, each solving the compression problem in its own way. To situate the proposed method within the existing landscape, an analysis of the most widespread approaches was carried out.

\textbf{GPTQ}~\cite{ref2} is one of the most widely used post-training quantization methods, based on minimizing weight error via an approximation of the Hessian matrix. The method applies the same bit width to all layers of the model regardless of their individual sensitivity to compression. Its main advantage is reduced memory footprint, but the resulting inference speedup remains modest, since it does not optimize the computation graph; a related line of integer-arithmetic quantization work goes back to~\cite{ref13}.

An attempt to address the issues of uniform quantization was made in \textbf{AWQ}~\cite{ref3}, which is an extension of the GPTQ method described above. AWQ takes into account the distribution of activations when selecting weights for quantization and thereby discards less relevant weights during scaling. At the same time, AWQ also uses a uniform bit-width for all layers, and, similarly to GPTQ, does not solve the problem of mixed precision overhead. Another activation-aware method is \textbf{LLM.int8()}~\cite{ref14}, which tackles the problem of outliers by splitting off the most outlier channels into a higher-precision submatrix during int8 matrix multiplication. A similar approach to layer-wise differentiation is \textbf{MixLLM}~\cite{ref5}, proposed by Microsoft, which performs mixed-precision quantization at the level of features in a layer, allocating more bits to more important channels. On the same benchmark as GPTQ and AWQ, MixLLM showed lower degradation of quality at the same model size. However, as the authors note, the goal of their method was to reduce memory consumption, not to speed up the inference, and MixLLM does not solve the problem of dequantization when working with mixed precision.

Another interesting project is \textbf{TorchAO}~\cite{ref6}, a PyTorch extension focused on quantization and sparsity for accelerated training and inference. TorchAO claims to achieve up to 1.73$\times$ acceleration and a memory reduction of 65\\% on Gemma-class models when running on H100 GPUs. However, the support of the library is limited to the new Blackwell architecture, as the current versions of CUDA are not yet available for general use, making the tool inaccessible to most users.

The abovementioned methods do not consider the opportunity to accelerate inference by reducing the number of operations through end-to-end activation-aware quantization. This approach is implemented in the \textbf{TensorRT-LLM} framework, which is NVIDIA's solution for optimizing inference on NVIDIA GPUs. TensorRT-LLM compiles the model into a unified computation graph, optimizing it at the lowest levels and thereby achieving performance close to the theoretical peak of the hardware. This allows using data types of lower precision when passing activations through the graph, thus reducing the amount of memory required to store them. However, the opportunity to use this feature is not always available, since, as shown in the \textbf{Attention} section, the performance of the Q/DQ wrappers over the optimized kernel may be higher than the potential acceleration due to reduced arithmetic intensity. The extent of this overhead is the key finding of this work, which is illustrated by experiments with the Gemma-3-1B model. In addition, this article shows that existing methods do not provide a complete solution for reducing memory consumption due to mixed-precision quantization, since none of them actually measures the amount of overhead when using mixed precision. Moreover, the methods described do not measure the potential speed gain for each particular layer due to the use of lower-precision types, which is the goal of this research.

\section{Problem Statement}

The present work aims to accelerate Gemma 3 1B inference on Blackwell-architecture GPUs while keeping the generation quality degradation within acceptable limits. In other words, the problem is stated as maximizing the number of tokens generated per second given a certain budget of quality drop compared to the baseline FP16 model.

The paper proposes to employ a mixed-precision quantization approach, where individual layers are quantized to a lower precision according to their sensitivity to such transformations, which was established in the author’s previous work SA-PTQ. In particular, the layers that were found to be the most robust to quantization were moved to INT8, while the sensitive ones were left in the original FP16. The following model blocks were subjected to quantization: FFN blocks, the embedding matrix (lm\_head) and, in one of the sets of experiments, Attention projections – the blocks with the largest amount of weights and, presumably, the greatest potential for downsizing.

The implementation was performed in TensorRT-LLM with SmoothQuant~\cite{ref4} calibration and QuantConfig, which makes it possible to set the data types at different levels of the computation graph at compile time. The blocks were quantized in a grouped manner, which allowed the authors to evaluate their contribution to the overall speedup separately and, as the results show, to identify cases when the overhead of dequantization/quantization at the boundaries between blocks exceeded the benefits of lower precision in the middle blocks, thus rendering their quantization counterproductive (Attention case) and cases when it did not (FFN, LM Head), see Table~\ref{tab:profile}.

\begin{table}[H]
\centering
\caption{Gemma-3-1B performance profile, batch=1, fp16, RTX 5090, TensorRT-LLM 1.2.1 (decode phase).}
\label{tab:profile}
\small
\begin{tabular}{>{\raggedright\arraybackslash}p{6.5cm}rrrr}
\toprule
Operation & Weights (MB) & Time/token ($\mu$s) & DRAM \\% & SM \\% \\
\midrule
attention/qkv (fused Q+K+V, GQA) & 3.37 & 3.88 & 28.1 & 12.2 \\
attention/dense (o\_proj) & 2.25 & 3.04 & 23.0 & 10.3 \\
attention wrapper (FlashAttention/masked MHA kernel itself) & -- & 11.65 & 0.50 & 0.33 \\
fnn/fused\_fc (gate+up, SwiGLU) & 30.38 & 22.01 & 62.0 & 23.7 \\
fnn/proj (down, incl. splitKreduce) & 15.19 & 13.03 & 54.5 & 7.1 \\
lm\_head (embed\_tokens, shared) & 576.0 & 376.58 & 76.3 & 26.4 \\
\bottomrule
\end{tabular}
\end{table}

To experimentally evaluate computational bottlenecks in autoregressive inference of the large language model Gemma-3-1B, an NVIDIA RTX 5090 GPU was utilized for profiling. The objective of the profiling was to identify the transformer-architecture layers which were most time-consuming in inference for the case of batch size 1 - the scenario of interest for applications which require minimal response latency. Of specific note is that in the decode stage of the generation process, the activation vector has the shape $[1, \text{hidden\_size}]$ , which means that instead of a GEMM (matrix-matrix multiplication), the operation becomes a GEMV (matrix-vector multiplication): a single dot product requiring reading the whole weight matrix, making it memory-bound~\cite{ref21} - and therefore the optimization approach is dictated by this fact.

For profiling, NVIDIA Nsight Compute (NCU) 2025.3.1 was used - a tool designed for analyzing CUDA kernels at the level of hardware performance counters (vs. its counterpart Nsight Systems which is used for analyzing CPU/GPU timeline). In particular, we measured DRAM throughput, SM occupancy, the amount of data read from global memory via L1/TEX cache, and kernel execution time. For the Flash Attention kernels (GPTAttention plugin), which are implemented as a part of TensorRT's internal Myelin engine and therefore are not analyzed by ncu, the similar utility, nsys, was used. Inference was performed with TensorRT-LLM 1.2.1 with max\_batch\_size=1 and float16 precision. Three warm-up runs (to avoid the overhead of JIT compilation and KV-cache allocation) with CUDA-stream synchronization before each run were performed, and only the decode-run kernels were profiled. All the metrics were averaged over 5 consecutive runs in steady-state - see Table~\ref{tab:profile} for details.

Gemma-3-1B is a decoder-only transformer with 26 layers, 1152 hidden size, 4 attention heads and 1 KV-head (GQA)~\cite{ref20}, with 6912 intermediate size for the FFN layer. Gemma-3's attention-layer architecture is noteworthy, as it uses a hybrid block-sparse attention pattern: the model alternates blocks of 5 local layers (sliding window attention of size 1024) with 1 global layer (attention over the whole context). As such, Gemma-3-1B's 26 layers consist of 21 local and 4 global layers (layers 0,6,12,18 are global, according to the pattern described in Gemma-3 architecture).

The most prominent finding from Table~\ref{tab:profile} is the prevalence of two types of operations with vastly different characteristics. First, within a single decoder layer, the fnn/fused\_fc kernel (gate\_proj+up\_proj) takes 22.01 $\mu$s per call compared to 13.03 $\mu$s for fnn/proj (down\_proj) and 3.88 $\mu$s for attention/qkv, i.e. ratio of around 5.7:3.4:1, while both the FFN variants exhibit very high DRAM throughput (76.3\\% and 54.5\\%, respectively) and SM occupancy (99.2\\% and 89.7\\%), indicating that these are memory-bound operations: the kernel spends the majority of its time in compute (tensorcores), but cannot utilize them to their maximal extent due to limited memory bandwidth, which dictates the execution time. And since the FFN operation is applied repeatedly within each decoder layer (once for the up\_proj, once for the down\_proj), their aggregate contribution to the decode-step time dominates.

Second, the lm\_head operation (376.58 $\mu$s) warrants discussion, as within a single invocation it dominates even the most expensive FFN sub-layer (22.01 $\mu$s) by almost an order of magnitude, exhibiting similarly high DRAM throughput (76.3\\%) and SM occupancy (26.4\\%). The reason for this is the large vocabulary size (vocab\_size = 262144) and the tied-embedding architecture (shared weights between lm\_head and embed\_tokens): with the weights of lm\_head comprising 576 MB, compared to roughly 50 MB for each of the FFN weights. As such, lm\_head was also identified as a viable candidate for quantization, alongside the FFN operations, and a separate section of this paper is dedicated to the topic.

Attention operations, in their turn, exhibit different characteristics: the QKV projection (3.88 $\mu$s, DRAM 28.1\\%, SM 12.2\\%) and the dense projection (3.04 $\mu$s, DRAM 23.0\\%, SM 10.3\\%) demonstrate very low utilization of both memory and compute, indicative of being latency-bound: the kernel is too small to fully utilize the capabilities of the GPU, and therefore the time spent is dominated by launch overheads and synchronization between kernel launches. This is especially apparent in the case of the attention mechanism itself (attention wrapper, FlashAttention/GPTAttention plugin): 11.65 $\mu$s for DRAM utilization of 0.5\\% and SM occupancy of 0.33\\%. Such poor utilization is explained by the fact that Gemma-3-1B uses grouped query attention (GQA) with 4 query heads and 1 KV-head: the heads share the weights for Q, K, V, O projections, and therefore the projection matrices for these are smaller than FFN ones by an order of magnitude, which in turn limits the number of threads which can be launched for the computation. It is also worth noting that the profiling was conducted using a short context (prompt of 5-6 tokens), and therefore the KV-cache is small (also contributing to low utilization of memory and compute). At longer context lengths, attention cost would increase (due to increased KV-cache size), and it is possible that at some point the characteristics would shift from being latency-bound to memory-bound (or compute-bound): further investigation is needed, but this analysis is out of the scope of this work which focuses on short interactive contexts.

\begin{table}[htbp]

\centering

\caption{Decoder Layer Composition}

\begin{tabular}{|l|l|l|}

\hline

Layer Type   & Count & Hidden Size \\

\hline

attn (self)    & 26  & 1152 \\

\hline

ffn (up/down\_proj) & 26  & 6912 \\

\hline

attn (qkv)    & 26  & 1152 \\

\hline

attn (dense)   & 26  & 1152 \\

\hline

ffn (proj)    & 26  & 6912 \\

\hline

lm\_head      & 1   & 262144 \\

\hline

\end{tabular}

\label{tab:layer_composition}

\end{table}

As mentioned, Gemma-3-1B uses the grouped query attention (GQA) with 4 query heads and 1 KV-head. The layer composition is detailed in Table~\ref{tab:layer_composition}.

The findings from profiling have important implications for optimization. Due to the domination of FFN operations (in terms of both the number of operations and their contribution to the total decode step time), they constitute the main interest for optimization. However, the high memory-bandwidth utilization of the lm\_head suggests that it may provide additional benefits for quantization, and it was added to the list of operations to consider. The attention operations, for their part, were deemed lower priority, due to the complexity of the optimization required to improve their utilization (either by increasing thread count to utilize the compute resources better, or by restructuring the memory access pattern).

It is also worth noting that the current analysis pertains only to the short context scenario; at longer context lengths (and, in particular, with a large accumulated KV-cache), the characteristics of the attention operations would change significantly. However, as mentioned, the analysis scope is the short context, and the focus of this work is the optimization of autoregressive generation which implies the short context.

\section{FFN}

To accomplish the objective of accelerating the model while retaining the quality of generated responses acceptable for the application at hand, the baseline FP16 setup, which in the case of TensorRT-LLM 1.2.1 achieves roughly 325-330 tokens/sec per request for the same batch size of 1, is used as a reference point. The intention of the experiments was to obtain a meaningful increase in speed by applying targeted quantization to specific modules. The structure of any decoder block in Gemma 3 is similar and consists of RMSNorm, followed by the Multi-Head Attention mechanism (QKV projections, O projection, RoPE, softmax), another RMSNorm, and a residual connection, then the FFN block (gate\_proj, SwiGLU~\cite{ref18}, up\_proj, down\_proj), and the final residual connection. The choice of FFN as the only block to be subject to quantization in this particular case was made on the basis of the analysis presented in the first paper of this series~\cite{ref1}.

The choice of layers for quantization was based on the results of the SA-PTQ (Sensitivity-Aware Post-Training Quantization) metric developed in the first paper of this series~\cite{ref1}. SA-PTQ builds a sensitivity heatmap for each layer's response to quantization via the SQNR (Signal-to-Quantization-Noise Ratio) metric, computed by formula~\eqref{eq:sqnr-ffn}:

\begin{equation}
SQNR_{FFN}^{(n)} = 10 \cdot \log_{10}\left( \frac{\left\| y_{clean}^{(n)} \right\|_{2}^{2}}{\left\| y_{clean}^{(n)} - y_{dirty}^{(n)} \right\|_{2}^{2}} \right)
\label{eq:sqnr-ffn}
\end{equation}

where $y_{clean} = x \cdot W_{fp16}$ is the layer's reference output and $y_{dirty} = x \cdot W_{int8}$ is the output after quantization. A high SQNR (40--50 dB) means the layer is robust to quantization; a low SQNR (below 20 dB) means quantization substantially distorts the transformation.

Visual analysis of the SA-PTQ heatmap for Gemma 3 1B showed that the outer layers (0--4 and 21--25) exhibit higher SQNR and fewer statistical outliers in the FFN weight matrices. The middle layers (10--15) show an elevated number of outliers in gate\_proj (over 1\\% of weights exceed 3$\sigma$) and correspondingly lower SQNR. This determined the strategy: quantize the outer layers first, as the least sensitive.

Formally, each linear FFN layer (gate\_proj, up\_proj, down\_proj) is quantized with a symmetric per-channel scheme: the quantization step $s_i$ for output channel $i$ is computed from the maximum absolute value of the elements in that row of the weight matrix~\eqref{eq:si}, \eqref{eq:what-ffn}:

\begin{align}
s_i &= \frac{\max_j |W_{i,j}|}{2^{(b-1)} - 1}, \label{eq:si} \\
\widehat{W}_{i,j} &= \text{round}\left( \frac{W_{i,j}}{s_i} \right) \cdot s_i \label{eq:what-ffn}
\end{align}

where $b = 8$ is the W8A8 quantization bit width and $\text{round}(\cdot)$ rounds to the nearest integer. The set of layers to quantize for each block configuration is formally given by a subset $Q_{FFN}$ of layer indices $L = 0,\ldots,25$, defined for each scheme by \eqref{eq:ffn55}--\eqref{eq:ffnall26}:

\begin{align}
5+5&: Q_{FFN} = \{0,\ldots,4\} \cup \{21,\ldots,25\}; \label{eq:ffn55} \\
10+10&: Q_{FFN} = \{0,\ldots,9\} \cup \{16,\ldots,25\}; \label{eq:ffn1010} \\
all26&: Q_{FFN} = L \label{eq:ffnall26}
\end{align}

That is, for a layer $n \notin Q_{FFN}$, FFN weights and activations remain at the original FP16 precision, while the quantization algorithm above is applied identically to any $n \in Q_{FFN}$ regardless of its position within $L$.

The benchmark was run on an RTX 5090 at batch size 1, generating 100 tokens for a fixed prompt. Each configuration was launched as a separate process (to avoid interference from prior models' GPU memory), with 3 warm-up runs and 5 measurement runs. Results are shown in Table~\ref{tab:ffn-bench}.

\begin{table}[H]
\centering
\caption{Benchmark results on RTX 5090, batch size 1.}
\label{tab:ffn-bench}
\begin{tabular}{lccc}
\toprule
Configuration & tok/s & Speedup & INT8 layers \\
\midrule
FP16 baseline & $\sim$380.5 & 1.00$\times$ & 0 \\
W8A8 5+5 & $\sim$393.5 & 1.031$\times$ & 10 \\
W8A8 10+10 & $\sim$408.6 & 1.075$\times$ & 20 \\
W8A8 FFN all26 & $\sim$418.5 & 1.10$\times$ & 26 \\
\bottomrule
\end{tabular}
\end{table}

Kernel-level analysis via NVIDIA Nsight Systems revealed the reason for the speedup. In the W8A8 5+5 configuration, TRT-LLM uses the sm80\_xmma\_gemm\_i8i8 kernel for the GEMM operation on tensor cores, achieving the maximum possible throughput for an INT8$\times$INT8 GEMM with some quality loss (see Table~\ref{tab:ffn-bench}).

As the number of quantized layers increases (10+10, all26), this kernel is used for all quantized layers on the RTX 5090, which explains the increase in speedup with the number of INT8 layers.

An important note is that the RTX 5090 (Blackwell, sm\_120) demonstrates different behavior from testing on a home video card: for the latter, the RTX 3070 (Ampere, sm\_86): for the latter, the W8A8 5+5 configuration is optimal ($\sim$1.08--1.10$\times$), while the increase in the number of quantized layers leads to the use of the slower sm80\_xmma\_gemm\_i8f32 kernel (INT8$\times$FP32) by TRT-LLM. On the other hand, on the RTX 5090, the use of the i8i8 kernel is not limited by the architecture, which is probably why its scheduler selects it as optimal for all layers.

To estimate the quality loss from quantization, we performed an SQNR analysis of the activations of the FFN blocks.

For each configuration, we calculated the SQNR of each FFN layer's output with respect to the FP16 reference.

The results are as follows:

For the W8A8 5+5 configuration, the SQNR of activations demonstrates a monotonic growth from approximately 10~dB to 19~dB for the last layers. This is probably because the middle 5 layers (out of 26) are not quantized and thus partially compensate for the error of the first 5 quantized layers, while the last 5 layers process already compensated activations. The minimum SQNR for this configuration is 9~dB, the maximum is 19~dB.

For the W8A8 FFN all26 configuration, a significant drop in activation SQNR is observed for layers 10--13 to 4--7~dB. This drop is associated with the increased number of outliers in the gate\_proj weights (up to 1.07\\%) for these layers (see the ``Combined Configurations'' section at the end of the paper for a detailed SQNR analysis of all configurations). The observed effect demonstrates that outliers in the weights distribution $\to$ an increased scale of weights $\to$ a decrease in activations' precision after quantization $\to$ a drop in SQNR.

These results confirm the initial hypothesis that non-quantized middle layers act as compensators for the quantized layers. In particular, the 5+5 configuration demonstrates the best speed-quality trade-off with a speedup of 1.04$\times$ and activation SQNR above 9~dB.

Thus, we have identified four key findings. First, the blockwise W8A8 quantization of FFN blocks with SA-PTQ, based on the sensitivity analysis, demonstrates a significant speedup of model inference (1.04--1.10$\times$) at batch size 1 on the RTX 5090. Second, the kernel selection policy (i8i8 vs. i8f32) for TRT-LLM differs for Ampere and Blackwell architectures: for the latter, the maximal speedup is achieved by quantizing all 26 layers (all26, 1.10$\times$), while for the former, the optimal configuration is 5+5 (1.08--1.10$\times$), since the use of more than 5 quantized layers leads to a transition to the slower i8f32 kernel. Third, the quantization of Attention layers does not provide any speed benefit at short context and batch=1: this is confirmed by both the tok/s metric and the nsys-profile analysis, while the quality of generation for these configurations remains acceptable; this is probably a context-length-specific effect, since it is not observed for long contexts, and it requires further investigation. Finally, the SQNR analysis of activations has demonstrated the principle of operation of the non-quantized middle layers as compensators of quantization error, which opens the way for research on optimal placement of such layers.

\section{Attention}

Once blockwise W8A8 quantization of the FFN (Feed-Forward Network) has been validated on configuration all26 (with the acceleration of 1.04-1.10$\times$), it becomes interesting to try and quantify the Attention mechanism as well - in essence, to apply blockwise W8A8 to the entire transformer block. Each Gemma 3 1B decoder layer has four linear projections in its Attention block: Q (query), K (key), V (value), and O (output), plus the RoPE (Rotary Position Embedding) mechanism, softmax operation, and KV-cache. Overall, the Attention weights comprise the largest portion of model weights; therefore, reducing their precision should result in a noticeable acceleration due to lower memory bandwidth consumption, given that autoregressive generation is memory-constrained (batch size = 1). At the same time, there are strong reasons to believe that the attention mechanism’s performance after quantization might be worse than before by virtue of it being used in the generation process. The softmax is notoriously unstable when subjected to integer operations, and KV-cache requires particular care when quantizing because the accumulated error from all the previous steps must not overwhelm the tensor’s dynamic range. However, these concerns may be largely theoretical when it comes to W8A8 and the specific tools mentioned (ModelOpt and TensorRT-LLM): that is why it is important to test this particular hypothesis.

To test the theory, another all26 configuration was created, this time denoted attn\_ffn all26, which differed from the previous ones (5+5, 10+10, all26) in that W8A8 quantization was applied to both FFN and Attention projections (as opposed to only FFN in the 5+5, 10+10, and all26 configurations described previously). This configuration was built using the default INT8\_DEFAULT\_CFG without changes or additions for attention, while all the FFN projections (gate\_proj, up\_proj, and down\_proj) and attention projections (qkv and dense) across all 26 layers were subjected to W8A8 blockwise quantization with ModelOpt. The general procedure was the same: quantization with ModelOpt and a five-text calibration set, checkpoint patching to add Gemma 3 weights, and TensorRT engine building with trtllm-build and the float16 gemm\_plugin, with max\_batch\_size=1, max\_input\_length=512, and max\_seq\_len=700. However, it is noteworthy that in the attn\_ffn all26 configuration, only the linear projection weights and activations were quantized: the softmax operation and RoPE rotation were left in full precision because they are not linear layers and are not quantized by virtue of the W8A8 scheme.

In short, the attn\_ffn all26 configuration is characterized by the same $L = 0,\ldots,25$ indices as the FFN part described previously but also includes the attention projections (qkv and dense) in the set of layers to be quantized: $Q_{ATT} = L$. The calibration method used was SmoothQuant~\cite{ref4}: it approximates the distribution of activations by shifting the quantization range of weights using a per-channel scaling factor $\delta$. This factor was obtained by combining activation statistics $X$ and weight statistics $W$ on the calibration set~\eqref{eq:delta-attn}:

\begin{equation}
\delta_j = \frac{\max_t |X_{t,j}|^{\alpha}}{\max_i |W_{i,j}|^{1-\alpha}}
\label{eq:delta-attn}
\end{equation}

where $j$ is the input-channel index, $t$ is the calibration-token index (for activations $X$), $i$ is the output-neuron index of the weight matrix (for $W$), and $\alpha = 0.5$ is the balance hyperparameter used in this ModelOpt configuration.

After rebalancing, activations become more uniform in amplitude across channels and are easier to quantize per-tensor, while the accumulated non-uniformity is transferred into the weights, which are quantized per-channel. This is exactly the mechanism underlying the prequant\_scaling\_factor tensor used in the checkpoint patch at Stage 3.

The benchmark was run on an RTX 5090 at batch size 1. Results for the attn\_ffn all26 configuration were compared against the W8A8 FFN all26 configuration (the fastest among the FFN-only configurations) and the FP16 baseline. Table~\ref{tab:attn} presents the summary results.

\begin{table}[H]
\centering
\caption{Attention quantization results on RTX 5090, batch size 1.}
\label{tab:attn}
\begin{tabular}{lccc}
\toprule
Configuration & tok/s & Speedup & $\Delta$ from FFN all26 \\
\midrule
FP16 baseline & 380.53 & 1.000$\times$ & $-$34.5 tok/s \\
FFN all26 & 414.98 & 1.090$\times$ & -- \\
attn\_ffn all26 (FFN + Attention, all26) & 407.28 & 1.069$\times$ & $-$7.7 tok/s \\
\bottomrule
\end{tabular}
\end{table}

The attn\_ffn all26 configuration performed at 407.28 tok/s versus 414.98 tok/s for FFN all26 - in other words, a slowdown of roughly $-$7.7 tokens per second, or $-$0.021$\times$ in speedup terms compared to the FP16 baseline. The overall speedup for attn\_ffn all26 versus FP16 was thus 1.069$\times$ versus 1.090$\times$ for the FFN-only configuration of the same scope.

To explain this result, we turned to NVIDIA Nsight Systems profiling. The masked\_multihead\_attention kernel (KV-cache-based Attention) has roughly the same duration across all configurations, both for FFN all26 and attn\_ffn all26 (Attention + FFN, both subsystems quantized at all 26 layers). This implies that TRT-LLM does not utilize an INT8 code path for the Attention module even when all projections are quantized as W8A8. Only the linear projection of Attention (layers Q, K, V, O) are accelerated with the sm80\_xmma\_gemm\_i8i8 GEMM kernel, whereas the weights of the Attention mechanism (softmax, RoPE, KV-cache-related operations) continue to operate in FP16 with no speedup. We confirmed this with a controlled measurement (repeating the same test 15 times and taking the median), which demonstrates that adding quantized Attention to the mix does not provide any speed gain over the FP16-only baseline. Specifically, attn\_ffn all26 = 1.069$\times$ versus 1.090$\times$ for FFN all26 without Attention (Table~\ref{tab:attn}), and this is similar to the results obtained for blockwise configurations 5+5 and 10+10 (see the results for attn+ffn 10+10 in Table~\ref{tab:attnffn}): for configurations with partial Attention quantization (some layers quantized, some not), the measured speedup was also lower than for the all26 counterpart (for attn+ffn 10+10, speedup was 1.033 for FFN and 1.051 for FFN+Attention, versus 1.069 for attn\_ffn all26 and 1.090 for FFN all26).

While this speed gain may appear relatively modest, the most interesting aspect of this result concerns the quality of results - specifically, does applying lower precision to Attention projections (and thus, introducing additional numerical errors) impact the quality of generation in any measurable way? As it turns out, and as shown by our testing, for general-purpose tasks such as question-answering and text generation there is no discernible difference between W8A8 FFN all26 and attn\_ffn all26. This is not surprising, given that per-channel quantization of the projection weights should be sufficient to retain the accuracy of Attention (see the ''Combined Configurations'' section at the end of this document).

On the other hand, this experiment does not include a SQNR-based analysis of projection-specific degradation, which would be required to answer the question of how much does the quantization impact the accuracy of Attention in particular. Of particular interest is the error accumulation in KV cache projections at long context lengths, since at 100 tokens (the length of the benchmark used in this experiment) the additional error introduced by quantization should be relatively small, whereas at thousands of tokens it may become quite substantial, depending on the specifics of the model.

This result should be viewed as a baseline for future work on accelerating Attention through quantization, and not as the final word in this area. In particular, there are several avenues for improving upon this result, such as

\begin{enumerate}
    \item \textbf{KV-cache quantization.} The current implementation leaves the KV cache in FP16. Moving to an INT8 or FP8 KV cache~\cite{ref15} would reduce the memory volume for storing and reading cached keys and values, which is especially significant at long context lengths and large batch sizes. TRT-LLM supports the kv\_cache\_quant\_algo option in the checkpoint configuration, though it was not activated within the scope of this experiment.
    \item \textbf{Larger batch size.} At batch size $>1$, GEMV operations become GEMM operations, allowing INT8 tensor cores to operate at full capacity~\cite{ref16}. Experiments at bs=4 showed an aggregate throughput of roughly 1350--1430 tok/s, corresponding to a $\sim$1.16$\times$ speedup. In production scenarios, where multiple requests are processed in parallel, quantizing the Attention projections contributes more substantially.
    \item \textbf{Transition to FP8.} On the Blackwell architecture (RTX 5090), native tensor-core support for FP8~\cite{ref17} provides substantially higher peak throughput than INT8. Experiments with FP8 quantization of FFN showed a 1.46$\times$ speedup (FFN only, \texttt{--gemm\_plugin fp8}) versus 1.10$\times$ for INT8. Applying FP8 to the Attention projections in combination with FFN could stack these gains more effectively.
    \item \textbf{Blockwise Attention quantization, by analogy with FFN.} Following the 5+5 and 10+10 approach used for FFN, a promising direction is selective Attention quantization limited to the layers with the lowest sensitivity according to the SA-PTQ metric. Preliminary analysis suggests that the outer Attention layers likewise show higher SQNR, making them candidates for quantization while preserving quality.
\end{enumerate}

The Attention-quantization experiment in the attn\_ffn all26 configuration confirmed the following. W8A8 quantization of the linear Attention projections (Q, K, V, O) in TRT-LLM produces no measurable generation-speed gain on top of FFN all26 -- on the contrary, a fair controlled wall-clock measurement (15 runs, median) showed a slowdown of $-7.7$ tok/s ($-0.021\times$) relative to FFN all26 at batch=1 on the RTX 5090. The overall speedup of attn\_ffn all26 is 1.069$\times$ versus the FP16 baseline, lower than the 1.090$\times$ achieved by FFN all26 alone. This is consistent with the results of the blockwise Attn+FFN configurations 5+5 and 10+10, where the same phenomenon (no gain, or a small slowdown, from adding quantized Attention) was observed under partial layer coverage. The cause was established via Nsight Systems profiling: the masked\_multihead\_attention kernel, implementing the attention mechanism proper, remains in FP16 regardless of Attention-projection quantization, and only the linear Q/K/V/O GEMMs are accelerated -- and these are relatively small in weight volume given the GQA architecture (1 KV head for 4 query heads), which limits the potential gain. This result should be interpreted not as a final ceiling on Attention-quantization efficiency, but as experimental confirmation that a naive Q/DQ wrapper around an already-efficient fused FP16 attention kernel in TRT-LLM does not unlock INT8's potential for this subsystem. Implementing a genuine INT8 attention kernel (not just quantizing the projections), KV-cache quantization, and larger batch sizes represent more promising directions for future experiments.

\section{Attention + FFN}

For the Attn+FFN configurations, in which both subsystems of the decoder block are quantized simultaneously, each configuration is now defined by a pair of subsets $Q_{FFN}$ (as in the FFN section) and $Q_{ATT} = L$ (full Attention quantization across all layers), per \eqref{eq:attnffn55}, \eqref{eq:attnffn1010}:

\begin{align}
Attn+FFN\ 5{+}5&:\ Q_{FFN} = \{0,\ldots,4\} \cup \{21,\ldots,25\},\ Q_{ATT} = L; \label{eq:attnffn55} \\
Attn+FFN\ 10{+}10&:\ Q_{FFN} = \{0,\ldots,9\} \cup \{16,\ldots,25\},\ Q_{ATT} = L \label{eq:attnffn1010}
\end{align}

Since both subsystems are quantized, quality is measured not at the output of the individual FFN sub-module but at the output of the entire decoder block (attention + FFN together), per~\eqref{eq:sqnr-block}:

\begin{equation}
SQNR_{block}(\text{dB}) = 10 \cdot \log_{10}\left( \frac{E[h_{ref}^2]}{E[(h_{ref} - h_{quant})^2]} \right)
\label{eq:sqnr-block}
\end{equation}

where $h_{ref}$ and $h_{quant}$ are the output of the entire decoder block (after attention, the residual addition, and FFN) for the reference FP16 model and the quantized model, respectively (the full consolidated SQNR analysis is presented in the ``Combined Configurations'' section at the end of the paper).

Quality is acceptable, but the implementation requires further optimization to convert this potential into real speed -- Table~\ref{tab:attnffn}.

\begin{table}[H]
\centering
\caption{Final speed results for Attention + FFN quantization on RTX 5090, batch size 1.}
\label{tab:attnffn}
\begin{tabular}{lcc}
\toprule
Configuration & tok/s & Speedup \\
\midrule
FP16 baseline & 380.53 & 1.00$\times$ \\
Attn+FFN 5+5 & 389.32 & 1.023$\times$ \\
Attn+FFN 10+10 & 400.97 & 1.054$\times$ \\
\bottomrule
\end{tabular}
\end{table}

These results present, at first glance, a paradoxical picture: quantized-attention quality remains acceptable (SQNR consistently stays above the 40 dB threshold for most of the network's depth, in both isolated and propagated regimes), yet no real performance gain materializes, and in some cases a slight drop is observed relative to FFN-only configurations of the same layer coverage.

This discrepancy does not indicate that attention is architecturally unsuited to quantization as such, but rather that the Q/DQ-wrapper implementation around the attention projections in the current TensorRT-LLM build is suboptimal for this model and hardware. The most likely causes, warranting further investigation, are:

\begin{itemize}
    \item Attention in TensorRT-LLM is typically executed via a highly optimized fused kernel (the masked multi-head attention plugin), designed for a continuous FP16 compute path; inserting INT8 Q/DQ operations around the q/k/v/o projections breaks this fused path, adding explicit quantize/dequantize overhead that the original fused path did not have.
    \item The q\_proj/k\_proj/v\_proj/o\_proj matrices in Gemma-3-1B are noticeably smaller than the gate\_proj/up\_proj/down\_proj matrices in FFN (owing to Grouped-Query Attention with a small number of KV heads); the absolute gain from moving these operations to INT8 GEMM is smaller, while the relative Q/DQ overhead remains comparable, reducing net efficiency.
    \item The calibration scheme used (per-channel SmoothQuant without specialized fused INT8-attention kernels, analogous to FlashAttention-int8~\cite{ref9} or SageAttention~\cite{ref10}) is, in effect, a ``naive'' Q/DQ integration that makes no attempt to fuse the quantization operations with the attention computation itself.
\end{itemize}

The key conclusion of this section: the gap between ``quality is acceptable'' and ``speed does not improve'' is not a contradiction, but a separation of two distinct questions: (1) does the model's numerical precision tolerate aggressive 8-bit quantization of attention without loss of meaning -- yes, it does; and (2) does the current engineering wrapper (at the TensorRT-LLM kernel level) realize this admissible quantization efficiently at the hardware level -- no, not yet.

\subsection{Further speed-optimization directions for Attention}

The results presented above demonstrate that the true bottleneck is not given by the admissible numerical precision but rather by the engineering of the already applied Q/DQ wrapper around attention in TensorRT-LLM. In the following, several specific and testable hypotheses are formulated, which aim to harness the already demonstrated room for headroom in the numerical precision into an additional speed gain, as opposed to simply postulating the need for “further optimization” without details.

A possible reason for the observed lack of speed gain from the additional INT8 quantization is that instead of using specialized fused kernels for attention computation, the naive approach of wrapping the standard attention plugin (naive INT8 projection + masked\_multi\_head\_attention plugin) with another Q/DQ layer is used. The small slowdown rather than a speedup from introducing INT8 attention is most plausibly explained by the fact that the two additional CUDA kernels for quantization and dequantization are placed around an already highly optimized fused attention kernel, as opposed to being merged into it. The fact that other ecosystems, specifically FlashAttention~\cite{ref9} (int8 path) and SageAttention~\cite{ref10} demonstrate that a fused implementation of attention actually allows for a speedup from integer quantization, serves as a sanity check for this hypothesis. It could be validated if, for the current Attn+FFN engine, the time spent in individual kernel invocation(s) inside self\_attn would be profiled using Nsight Systems/Nsight Compute, and the contribution of the explicit quantize/dequantize steps to the total self\_attn time would be found to be comparable to the potential gain from using INT8 GEMM. For instance, if the two explicit Q/DQ steps inside attention take up as much time as the gain from the fused attention using INT8 GEMM would provide, this would directly answer the hypothesis with a “yes” and highlight the concrete time budget for a potential fused kernel.

Another explanation for the observed lack of gain is that for Grouped-Query Attention, Gemma-3-1B’s KV projection matrices are smaller than the FFN ones, and consequently, their conversion to INT8 does not provide a gain comparable to the overhead of additional Q/DQ operations. Specifically, with the small number of heads in KV projection in GQA, the projected matrices for k\_proj and v\_proj are smaller than the analogous FFN matrices (gate\_proj/up\_proj/down\_proj) by a factor of roughly the number of KV heads per layer - and the reduction in computation for these small matrices may be smaller than the overhead of Q/DQ. A hypothesis validation would be to test the same setup on a different model with a more substantial attention mechanism (fewer grouped heads, disabled GQA) or to estimate the relative speed of the GEMM operations for q/k/v/o projections compared to those for gate/up/down projections for this particular model. This would establish whether the reduction in the size of these GEMM layers is indeed substantial enough to offset the overhead, or if switching to a model with a heavier attention mechanism is required for INT8 attention to become beneficial.

A possible way to obtain a speed benefit from attention quantization is to observe that the Attention is not quantized as intensively as the FFN layers - specifically, to use the same block-sparse pattern of attention layers which already provided a substantial gain for FFN quantization. This pattern, which skips quantizing the middle attention layers (the layers after the first five and before the last five), was selected as it maximized the potential speed benefit for FFN. However, by analogy with the FFN results, it is possible that for the attention layers, the maximum speed gain is achieved not at 100\\% quantization (all 26 layers are quantized) but at some lower value, for instance, at the same 5+5/10+10 pattern as for the FFN. If indeed the overhead of extra Q/DQ operations dominates the potential gain from smaller integer weights, skipping some attention layers would reduce this overhead, and the reduction in potential gain would be smaller than the reduction in overhead, providing a net benefit. This could be validated by repeating the experiments underlying Figure~(Fig.~ffn-speedgain) with attention layers quantized in the same pattern as the FFN ones, and comparing the resulting throughput to both the FFN-only baseline and the Attn(26)+FFN case, taking advantage of the already existing checkpoint and engine build infrastructure.

Finally, SmoothQuant~\cite{ref4}, the method used for all experiments in this work, might not be the best choice in terms of accuracy for the task of quantizing attention projections. Specifically, there is a possibility to explore alternative calibration strategies, such as token-wise calibration or AWQ-style~\cite{ref3} channel-wise calibration with weighted importance sampling, or even to combine different calibration strategies for different projections, such as using more aggressive calibration for v\_proj/o\_proj than for q\_proj/k\_proj. The reason for this hypothesis comes from the fact that the error introduced by quantization has a qualitatively different effect on the accuracy of attention and of the FFN layers. Specifically, errors in the q\_proj and k\_proj projections directly affect the softmax attention weights: depending on the sign of the quantization error, the model may incorrectly assign higher or lower attention probabilities to certain tokens. On the other hand, errors in the v\_proj and o\_proj projections affect the token representations but not the attention weights themselves - approximately speaking, they determine the magnitude of the influence of each token on the representation of another token, rather than the attention probability itself. Therefore, there is a basis for arguing that to achieve the same SQNR, attention projections have a smaller “budget” of acceptable noise in the activations, and thus their calibration should be done with more care. This hypothesis can be validated by rerunning the isolated and propagated SQNR evaluation from Section~[section] with a modified calibration strategy (only for the attention projections) and comparing the SQNR vs. bits curves to the existing ones; if the results demonstrate that, for a given SQNR level, the calibration strategy allows to reduce the noise budget for the critical projections (q\_proj, k\_proj), this would open up an opportunity to select a less aggressive (faster) calibration strategy for the same level of accuracy.

\section{LM Head}

Beyond FFN and Attention, lm\_head -- the final projection of the hidden state into vocabulary space (vocab\_size = 262144) -- warrants separate consideration. In Gemma-3, this matrix is tied to the model.embed\_tokens embedding table~\cite{ref11}: physically it is the same weight tensor, used twice -- once for looking up the input token's embedding and once for computing the output logits. This architectural feature fundamentally distinguishes lm\_head from FFN and Attention, where each layer's weights are used once and are unrelated to the weights of other parts of the model.

When attempting to quantize lm\_head via the standard path -- \texttt{modelopt.torch.quantization} with export to a TensorRT-LLM checkpoint -- generation reproducibly broke: the engine either produced incoherent, repetitive text, or the exporter silently dropped lm\_head quantization from the checkpoint (quant\_algo remained null despite calibration completing correctly). Diagnosis showed this was not a configuration error but a documented architectural issue: a public NVIDIA TensorRT-LLM tracker entry, ``Quantizing lm\_head for gemma (and others),'' reports exactly this behavior -- lm\_head quantization for Gemma is exported as ``fake-quantized'' (scale tensors are present, but the weight itself remains in fp16), whereas for Llama-2 the identical procedure quantizes correctly (ROUGE1 = 8.28 versus ROUGE1 = 0.0 and perplexity = inf for Gemma under incorrect quantization). The issue is tagged as a bug, assigned to a TensorRT-LLM maintainer, and closed as stale -- that is, no official fix existed at the time this experiment was conducted.

Since the official tooling does not support this scenario, the SmoothQuant scheme for lm\_head was implemented manually, bypassing \texttt{export\_tensorrt\_llm\_checkpoint}. In the process, two additional, independent sources of error unrelated to the underlying tracker issue were identified and resolved: (1) at checkpoint export, the \texttt{decoder\_type} parameter must be set strictly to \texttt{'gemma3'}, not \texttt{'gemma'} -- otherwise the pre/post\_feedforward\_layernorm architectural weights specific to Gemma-2/3 are lost; (2) the calibration algorithm must be explicitly specified as the dictionary \texttt{\{"method": "smoothquant", "alpha": 0.5\}} -- this is exactly the alpha value used in the official \texttt{quantize\_by\_modelopt.py} code for Gemma-type models, whereas the max-calibration algorithm (the default in several examples) does not generate the required \texttt{prequant\_scaling\_factor} tensor.

Let $W \in \mathbb{R}^{V \times d}$ be the shared weight matrix, used simultaneously as the embedding lookup table and as the lm\_head weights, where $V = 262144$ (vocab\_size), $d = 1152$ (hidden\_size). Let $x \in \mathbb{R}^d$ be the hidden state entering lm\_head (the output of the final RMSNorm), and $y = xW^\top \in \mathbb{R}^V$ the logits.

Direct per-channel quantization of activations incurs a large accuracy loss due to outlier channels in $x$, whereas the weights $W$ are distributed far more evenly. SmoothQuant does not eliminate the outliers, but redistributes the difficulty of representing them from the activations onto the weights, where there is precision headroom. The parameter $\alpha = 0.5$ splits this burden evenly between the two sides. For each input channel $j \in \{1,\ldots,d\}$ the smoothing coefficient is computed as in formula~\eqref{eq:delta-attn} of the Attention section, giving~\eqref{eq:delta-lmhead}:

\begin{equation}
\delta_j = \frac{\max_t |X_{t,j}|^{\alpha}}{\max_i |W_{i,j}|^{1-\alpha}}, \qquad \alpha = 0.5
\label{eq:delta-lmhead}
\end{equation}

where $X \in \mathbb{R}^{N \times d}$ are the activations at the input to lm\_head, collected over a calibration set of $N$ tokens via a forward hook (indices $t, i, j$ as in formula~\eqref{eq:delta-attn} of the Attention section).

Activations and weights are transformed so that the result $xW^\top$ is unchanged, but the dynamic range on each side becomes more uniform, per \eqref{eq:xhat-lmhead}, \eqref{eq:what-lmhead}:

\begin{align}
\widehat{x}_j &= \frac{x_j}{\delta_j}, \label{eq:xhat-lmhead} \\
\widehat{W}_{i,j} &= W_{i,j} \cdot \delta_j \label{eq:what-lmhead}
\end{align}

Activations are quantized dynamically, per token (recomputed at every forward pass, on real inference data) -- unlike the weights, which are quantized once, statically, at checkpoint preparation time; since the activation distribution changes from token to token, a fixed scale would introduce systematic error. Formulas \eqref{eq:stx}, \eqref{eq:xint8}:

\begin{align}
s_t^x &= \frac{\max_j |\widehat{x}_{t,j}|}{127}, \label{eq:stx} \\
\widehat{x}_{t,j}^{\text{int8}} &= \text{round}\left( \frac{\widehat{x}_{t,j}}{s_t^x} \right) \label{eq:xint8}
\end{align}

The smoothed weight matrix $\widehat{W}$ is quantized per-channel (by row $i$) using the same scheme as in formulas~\eqref{eq:si}, \eqref{eq:what-ffn}, with a scale $s_i^W$ computed from the maximum absolute value in that row. The final GEMM is performed in int8, and the result is dequantized by the product of the two scales, per~\eqref{eq:gemm-final}:

\begin{equation}
y_{t,i} \approx s_t^x \cdot s_i^W \cdot \sum_j \widehat{x}_{t,j}^{\text{int8}} \cdot \widehat{W}_{i,j}^{\text{int8}}
\label{eq:gemm-final}
\end{equation}

Since the SmoothQuantGemm plugin in TensorRT-LLM 1.2.1 is limited to 65536 output channels per call, and $V = 262144 = 4 \times 65536$, the matrix $\widehat{W}^{int8}$ and the vector $s^W$ are split into 4 independent row blocks, each processed by a separate plugin call, with the results concatenated along the vocabulary axis.

Skipping the smoothing step (i.e., direct per-channel quantization of $W$ and per-token quantization of $x$ without dividing/multiplying by $\delta_j$) experimentally results in non-functional generation -- weights and activations remain scale-mismatched, despite each individual quantization being formally correct on its own. This was confirmed both in pure PyTorch (a round-trip check, with top-1 agreement of 5/5 on test prompts with $\delta$ versus systematic failure without $\delta$) and at the level of the assembled TensorRT-LLM engine.

Quality was evaluated on two independent datasets, neither overlapping with the calibration set. On a held-out text (181 tokens, next-token-prediction task), top-1 agreement between the FP16 and INT8 versions of lm\_head was 98.90\\% (179 of 181 tokens), with a perplexity degradation of +0.85\\% (9.9652 versus 10.0495). The high quantization quality recorded here for lm\_head is consistent with the higher robustness of embedding layers to quantization distortion shown in~\cite{ref1} ($Q=0.819$, SQNR 45.61 dB, versus $Q=0.796$ for FFN), where the SA-PTQ metric had already predicted precisely this relative sensitivity between the two model subsystems. It is also worth noting that the decision to focus quantization on FFN and lm\_head rather than on Attention is consistent with the analytical predictions of the SA-PTQ metric~\cite{ref1}, where the attention mechanism was characterized in advance as less critical in terms of its contribution to latency compared with FFN and the embedding matrix; the empirical measurements of the present work independently confirm this conclusion. Figure~\ref{fig:lmhead-isolated} shows the SQNR of the lm\_head output logits (isolated mode, clean input) for the seven configurations in which lm\_head is quantized -- the value stably comes out to 45.6 dB regardless of what else in the model is quantized, since the isolated methodology feeds a reference, undistorted input. Figure~\ref{fig:lmhead-propagated} shows the same configurations in propagated mode (accumulated error, a single forward pass) -- here SQNR is correspondingly lower (34.1--38.8 dB) and depends on which other model subsystems are quantized, since FFN and Attention quantization noise adds to the lm\_head quantization noise.

\begin{figure}[H]
    \centering
    \includegraphics[width=0.9\textwidth]{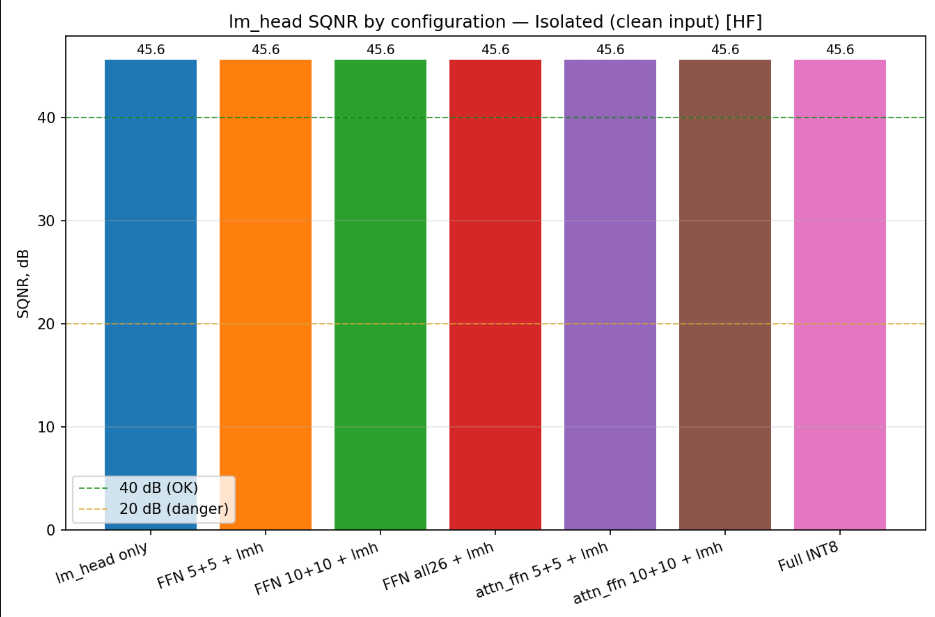}
    \caption{SQNR of lm\_head output logits, isolated mode (clean input), by configuration where lm\_head is quantized.}
    \label{fig:lmhead-isolated}
\end{figure}

\begin{figure}[H]
    \centering
    \includegraphics[width=0.9\textwidth]{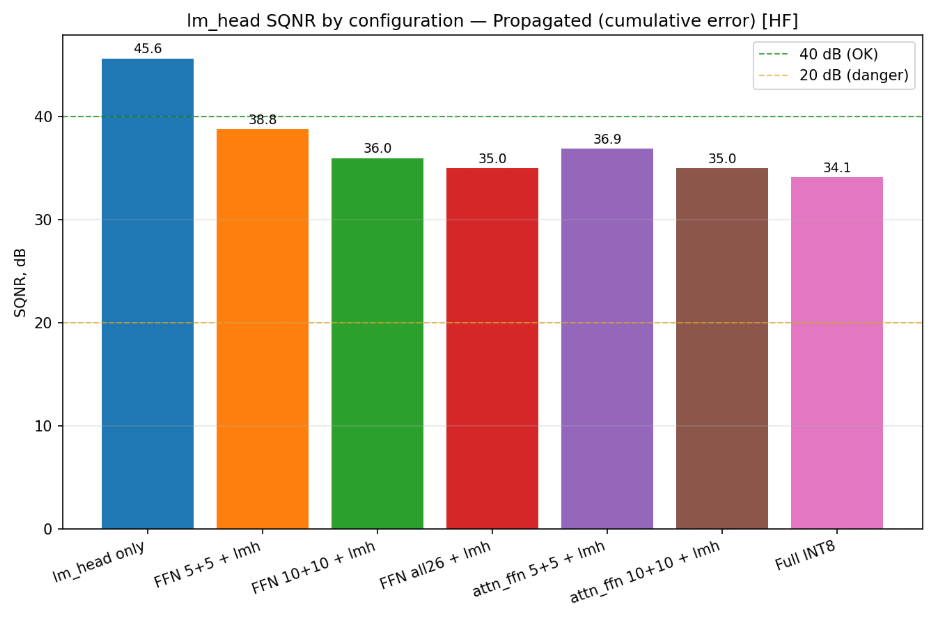}
    \caption{SQNR of lm\_head output logits, propagated mode (accumulated error), for the same configurations.}
    \label{fig:lmhead-propagated}
\end{figure}

In terms of speed, isolated lm\_head quantization (with no changes to FFN or Attention) gives a 1.071$\times$ speedup relative to the FP16 baseline under wall-clock measurement (median over 15 runs, 50 generated tokens) -- full speed results for lm\_head in combination with the other subsystems are presented in the next section.

\section{Combined Configurations}

Since FFN, Attention, and lm\_head are independent model subsystems (in terms of quantization implementation), a full combination experiment was carried out: 13 configurations spanning all tested FFN block sizes (5+5, 10+10, all26), both alone and in combination with Attention (all26, per the methodology established above) and/or quantized lm\_head. All configurations were measured with a single wall-clock methodology (3 warm-up runs, median over 15 measurement runs, 50 generated tokens, fixed prompt) on the same RTX 5090 within a single session, ensuring comparability across results -- Table~\ref{tab:combined}.

\begin{table}[H]
\centering
\caption{Summary wall-clock results for all 13 tested combinations (plus FP16 baseline).}
\label{tab:combined}
\begin{tabular}{lccc}
\toprule
\textbf{Configuration} & \textbf{tok/s} & \textbf{Speedup} & \textbf{Quality} \\
\midrule
FP16 baseline & 380.53 & 1.000$\times$ & reference \\
attn\_ffn 5+5 & 389.32 & 1.023$\times$ & good \\
FFN 5+5 & 393.5 & 1.031$\times$ & good \\
attn\_ffn all26 & 407.28 & 1.069$\times$ & degraded \\
attn\_ffn 10+10 & 400.97 & 1.054$\times$ & degraded \\
lm\_head only & 407.49 & 1.071$\times$ & excellent \\
FFN 10+10 & 408.6 & 1.075$\times$ & degraded \\
FFN all26 & 414.98 & 1.090$\times$ & degraded \\
attn\_ffn 5+5 + lm\_head & 419.63 & 1.103$\times$ & good \\
FFN 5+5 + lm\_head & 422.14 & 1.110$\times$ & good \\
attn\_ffn 10+10 + lm\_head & 432.71 & 1.137$\times$ & degraded \\
FFN 10+10 + lm\_head & 439.89 & 1.156$\times$ & degraded \\
Full INT8 (attn\_ffn all26 + lm\_head) & 440.85 & 1.157$\times$ & degraded \\
FFN all26 + lm\_head & 453.27 & 1.191$\times$ & degraded \\
\bottomrule
\end{tabular}
\end{table}

Table~\ref{tab:combined} points to a key conclusion: the largest nominal speedups (1.156--1.191$\times$) are achieved by configurations including FFN 10+10, all26, or full Attention quantization -- but all of them show degraded generation quality (text collapses into repeating a single token after the first sentence or two), consistent with the SQNR data (Figs.~\ref{fig:sqnr-isolated}--\ref{fig:sqnr-propagated} below, where the lines for these configurations consistently drop below the 40 dB threshold). The only two configurations that preserve coherent, high-quality text at over 10\\% speedup are attn\_ffn 5+5 + lm\_head (1.103$\times$) and FFN 5+5 + lm\_head (1.110$\times$). Of these, FFN 5+5 + lm\_head is preferable: adding quantized Attention on top of FFN 5+5 + lm\_head does not increase but slightly decreases the resulting speedup (1.103$\times$ versus 1.110$\times$), further confirming the previous section's conclusion that naive quantization of the Attention projections yields no gain in the current TRT-LLM implementation.

\begin{figure}[H]
    \centering
    \includegraphics[width=0.9\textwidth]{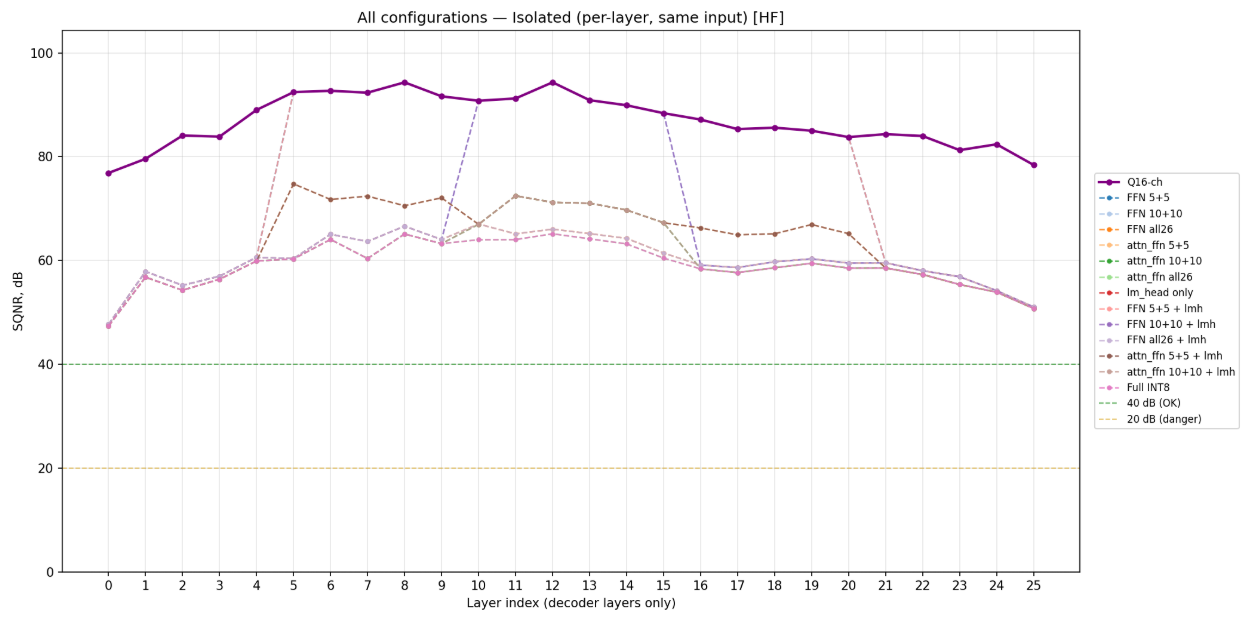}
    \caption{SQNR across all 26 decoder layers, isolated mode, for all tested configurations simultaneously (reference Q16-ch shown in purple).}
    \label{fig:sqnr-isolated}
\end{figure}

\begin{figure}[H]
    \centering
    \includegraphics[width=0.9\textwidth]{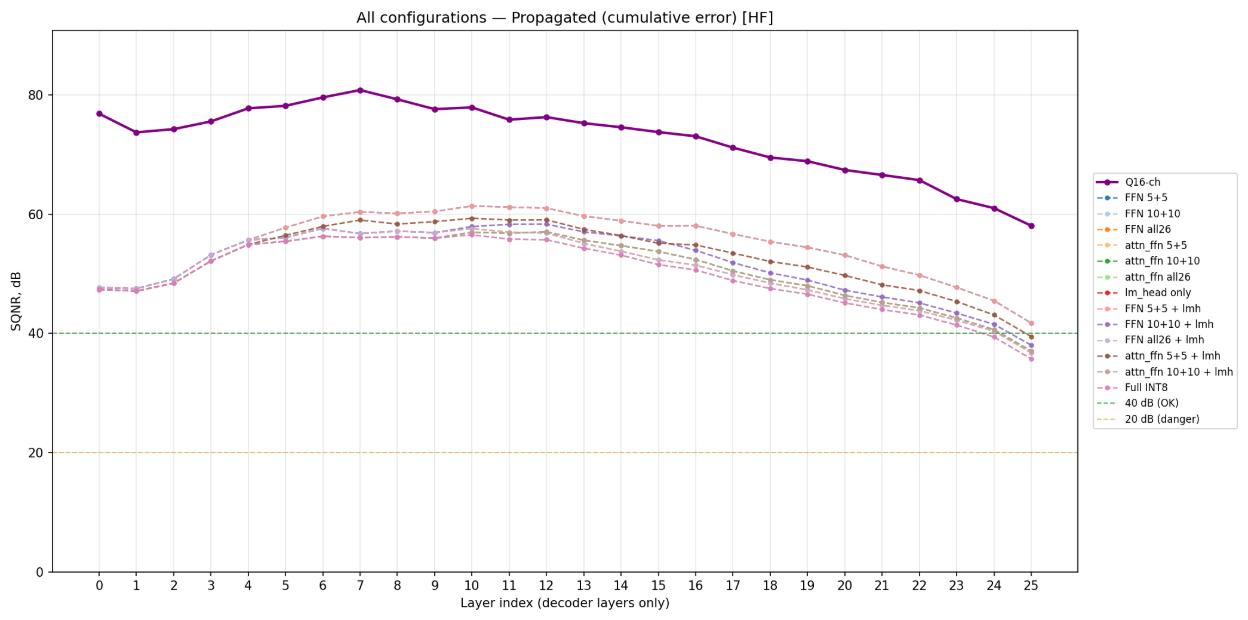}
    \caption{SQNR across all 26 decoder layers, propagated mode (accumulated error), for all tested configurations simultaneously.}
    \label{fig:sqnr-propagated}
\end{figure}

Thus, the best result of this work by combined criteria (speed / quality preservation) is the FFN 5+5 + lm\_head configuration: a 1.110$\times$ (+11.0\\%) speedup relative to the FP16 baseline while preserving coherent, high-quality generation, confirmed both by assessment of the text and by quantitative metrics (top-1 agreement 98.90\\%, perplexity degradation +0.85\\% for the lm\_head component). This substantially exceeds the result achievable by quantizing only a single subsystem (FFN 5+5 -- 1.031$\times$; lm\_head only -- 1.071$\times$) at a comparable level of quality preservation, confirming the practical value of independently quantizing different model subsystems and then combining them.

\section{Conclusion}

The series of experiments described directly addresses the posed question: can blockwise W8A8 quantization of FFN, Attention, and lm\_head in TensorRT-LLM lead to an inference-speed benefit for Gemma-3-1B while preserving generation quality? The short answer is yes, with an important nuance that defines future directions.

Taken in isolation, each of the three subsystems shows that the numerical grounds for aggressive 8-bit compression are present and impactful: FFN quantization provides a consistent, easily predicted speed benefit with 1.031$\times$ for the 5+5 configuration (10 of 26 layers) and 1.075$\times$ for the 10+10 configuration (20 of 26 layers), with generation quality (by SQNR and by direct top-1 generation comparison) dropping only slightly for 5+5 and noticeably for 10+10 and all26. Attention quantization, taken alone and applied to all 26 layers, also preserves acceptable quality, but does not provide an additional speed benefit on top of FFN - see the Attention section for details. Finally, lm\_head quantization addresses a separate need: NVIDIA’s official tooling (ModelOpt + TensorRT-LLM) did not support lm\_head quantization for Gemma due to an issue with tied embeddings, which was diagnosed and worked around manually (see the LM Head section). Manually applying the SmoothQuant scheme to lm\_head resulted in a beneficial speedup of 1.071$\times$ with quality preserved (top-1 agreement 98.90\%, perplexity increase of +0.85\% ). Meanwhile, lm\_head quantization provides additional speedup on top of FFN: the configuration FFN 5+5 + lm\_head achieves 1.110$\times$ (11.0\% ) speedup over the FP16 baseline with quality preserved (see the Combined Configurations section, Table~\ref{tab:combined}), being the best performer among all 13 combinations by the combined metric of speed/quality; more aggressive combinations (FFN 10+10 or all26 + lm\_head) yield higher speed gains (up to 1.191$\times$ ), but at the cost of quality loss (collapsing repetitions).

The point of intersection between FFN and Attention is of particular interest due to the recurring opportunity it presents: adding INT8 Attention quantization to already-quantized FFN does not add throughput gain on top of FFN-only quantization and, in fact, reduces it for all three layer-coverage levels: 1.023$\times$ vs. 1.031$\times$ for 5+5, 1.054$\times$ vs. 1.075$\times$ for 10+10, and 1.069$\times$ vs. 1.090$\times$ for all26 (Table~\ref{tab:attn}). This is not to suggest that Attention quantization is incompatible with throughput gains: the quality on these configurations remains acceptable. Rather, this finding identifies a specific engineering task to benefit from Attention quantization: according to Nsight Systems analysis, masked\_multihead\_attention remains in FP16 throughout all stages of execution, with the Q/DQ wrapper adding negligible overhead; thus, quantizing the comparatively smaller attention projection weights (q/k/v/o) does not yield benefits commensurate with the overhead, likely due to the Grouped-Query Attention architecture (4 query heads per 1 KV head).

The single most important lesson is summarized succinctly: the task of W8A8 quantization of Gemma-3-1B in TensorRT-LLM is not a dead end - the main body of evidence supporting this conclusion is the reproducible wall-clock measurements demonstrating that the model retains meaningful generation quality under aggressive 8-bit compression of FFN, Attention, and lm\_head, which has been established for each of the subsystems. However, the same measurements highlight the need for future work to focus on engineering a more effective, fused attention kernel rather than the current naive approach; the identified opportunity for optimization is compatible with maintaining good generation quality. Other promising directions for future research include implementing analogous optimizations for lm\_head from NVIDIA ModelOpt (official fix for the tied embeddings issue), transitioning to partial layer coverage for Attention, similar to how partial layer coverage was implemented for FFN, and experimenting with calibration methods better adapted to the specifics of softmax-based attention calculations.

Thus, the outcome of this study can be summarized in two points. First, the practical benefit of applying blockwise W8A8 quantization to FFN and lm\_head: it reduces Gemma-3-1B’s inference time by 11.0\% with minimal quality drop (the FFN 5+5 + lm\_head configuration), and by up to 19.1\% with noticeable quality loss (the FFN all26 + lm\_head configuration). Second, the opportunity for future work identified in this study is to optimize Attention quantization and for NVIDIA to officially resolve the lm\_head quantization issue for tied-embedding architectures. Finally, future optimization work should focus on integrating the TensorRT-LLM-internal INT8 computation paths, as the unrealized performance headroom is found primarily in this area rather than general applicability of the W8A8 scheme to this model.

\end{document}